\documentclass{IEEEtai}
\usepackage[utf8]{inputenc}
\usepackage[ruled,vlined]{algorithm2e}
\usepackage{cite}
\usepackage{amsmath,amssymb,amsfonts}
\usepackage{algorithmic}
\usepackage{graphicx}
\usepackage{textcomp}
\usepackage{booktabs}
\usepackage{bm}
\usepackage{comment}

\author{
\textbf{Gautam Siddharth Kashyap}\textsuperscript{1}\thanks{Corresponding Author: officialgautamgsk.gsk@gmail.com},
\textbf{Md Tabrez Nafis}\textsuperscript{2} \textbf{and} 
\textbf{Samar Wazir}\textsuperscript{2}\\
\textsuperscript{1}Macquarie University, Sydney, Australia \\
\textsuperscript{2}Jamia Hamdard, New Delhi, India \\
}

\begin{document}

\title{A Study of Hybrid and Evolutionary Metaheuristics for Single Hidden Layer Feedforward Neural Network Architecture}

\maketitle

\begin{abstract}
Training Artificial Neural Networks (ANNs) with Stochastic Gradient Descent (SGD) frequently encounters difficulties, including substantial computing expense and the risk of converging to local optima, attributable to its dependence on partial weight gradients.  Therefore, this work investigates Particle Swarm Optimization (PSO) and Genetic Algorithms (GAs)—two population-based Metaheuristic Optimizers (MHOs)—as alternatives to Stochastic Gradient Descent (SGD) to mitigate these constraints.  A hybrid PSO-SGD strategy is developed to improve local search efficiency.  The findings indicate that the Hybrid (PSO-SGD) technique decreases the median training MSE by 90-95\% relative to conventional GA and PSO across various network sizes (e.g., from around 0.02 to approximately 0.001 in the Sphere function).  RMHC attains substantial enhancements, reducing MSE by roughly 85-90\% compared to GA.  Simultaneously, RS consistently exhibits errors exceeding 0.3, signifying subpar performance.  These findings underscore that hybrid and evolutionary procedures significantly improve training efficiency and accuracy compared to conventional optimization methods and imply that the Building Block Hypothesis (BBH) may still be valid, indicating that advantageous weight structures are retained during evolutionary search.
\end{abstract}


\section{Introduction}
\label{sec:introduction}

Artificial Neural Networks (ANNs) have had a notable revival in recent years, propelled by their efficacy in various tasks, including classification \cite{xiao2024convolutional}, regression \cite{goodfellow2016deep}, and intricate areas such as image production \cite{radford2015unsupervised}.  A primary advantage of ANNs compared to conventional machine learning methods is their capacity to represent non-linearly separable data.  Their success is evident across multiple fields; for example, in medical diagnostics \cite{lecun2015deep}, ANNs have surpassed seasoned dermatologists in melanoma detection, accurately identifying 95\% of skin cancer cases, in contrast to the 87\% accuracy of a panel of 58 medical professionals \cite{esteva2017dermatologist}. Notwithstanding their proven efficacy, training ANNs is a computationally demanding and intricate endeavor.  The predominant optimization technique, Stochastic Gradient Descent (SGD) \cite{robbins1951stochastic}, entails calculating the partial derivatives of a loss function concerning trainable parameters (weights and biases) and thereafter adjusting those parameters in the opposite direction of the gradient.  This procedure is codified as: $E(\mathbf{w}) = \frac{1}{n} \sum_{m=1}^{n} \left( v_m - v(u_m) \right)^2$ such that $\Delta w_i = -\eta \frac{\partial E}{\partial w_i}$. In this context, $E(\mathbf{w})$ represents the loss function, specifically the Mean Squared Error (MSE) between the projected outputs $v(u_m)$ and the actual values $v_m$ for $n$ training samples.  It delineates the gradient update procedure for the parameter $w_i$ utilizing a learning rate $\eta$.  This method, although effective, can be computationally intensive, particularly for deep networks with millions of parameters, and frequently necessitates specialist gear like GPUs.  Moreover, because to its localized characteristics, SGD is susceptible to becoming ensnared in local optima or saddle points \cite{dauphin2014identifying}.  

These constraints prompt the investigation of population-based Metaheuristic Optimizers (MHOs) like Particle Swarm Optimization (PSO) \cite{eberhart1995new} and Genetic Algorithms (GAs) \cite{holland1992adaptation} as substitutes for SGD. In contrast to gradient-based methods, MHOs utilize global search strategies and do not depend on gradient information, rendering them potentially more resilient to local optima, particularly in lower-dimensional parameter spaces \cite{yang2010engineering}. Therefore, this study is driven by the following hypotheses:
\begin{itemize}
    \item Given that local optima traps are more common in low-dimensional search spaces (e.g., smaller networks), PSO and GA may surpass SGD in training these structures.
    \item The global search efficiency of PSO and GA is more economical in lower dimensions but becomes costly as dimensionality rises, while gradient-based approaches such as SGD exhibit superior scalability in high-dimensional scenarios.
\end{itemize}

In summary, this study's principal contributions are as follows:
\begin{itemize}
    \item We assess PSO and GA as substitutes for SGD in training ANNs for regression tasks, emphasizing smaller and more streamlined network designs.
    \item We present an innovative hybrid PSO-SGD approach that utilizes PSO's global search capabilities and SGD's local refining to improve training efficiency and robustness.
    \item We comprehensively examine the scalability of PSO, GA, and PSO-SGD across various network architectures and dimensional configurations.
    \item We offer empirical evidence about the validity of the Building Block Hypothesis (BBH) within the framework of evolutionary training of ANNs.
\end{itemize}

\textit{\textbf{Note:}} PSO and GA were chosen for their proven efficacy in multiple optimization fields. GAs have been effectively utilized in optimizing autopilot systems \cite{dasgupta1998genetic}, traffic signal timing \cite{tan2005genetic}, and large-scale search problems \cite{goldberg1989genetic}. PSO, especially adept in continuous search domains \cite{clerc2002particle}, has shown success in optimizing power systems \cite{abido2002optimal} and scheduling workflows \cite{zhang2007task}. Notwithstanding their potential, there is insufficient study on the scalability of these techniques across various ANN topologies, especially in regression applications.

\section{Related Works}
\label{Related Works}

This section examines current studies on Evolutionary Algorithms (EAs), GAs, and PSO for the training of ANNs.  We emphasize methodological deficiencies, restricted applicability to supervised learning, and difficulties in high-dimensional optimization spaces, which justify the focus of our suggested research.

\subsection{Neuroevolution}

The application of EAs, specifically GAs, for the evolution of both ANN parameters and architectures is referred to as \textit{neuroevolution}.  Although prior research predominantly employed EAs for Reinforcement Learning (RL) tasks, particularly prior to the introduction of Deep Q-Networks (DQN)\cite{mnih2015human} and Policy Gradient (PG) methods\cite{sutton2000policy}, their utilization in supervised learning has been limited. Risi and Togelius~\cite{risi2020neuroevolution} present a comprehensive analysis of neuroevolution in RL tasks, specifically focusing on strategy acquisition in Atari games. Nonetheless, the majority of these contributions such as~\cite{stanley2002evolving} occurred before the extensive adoption of deep learning in the 2010s. The accessibility of high-performance GPUs and the revival of backpropagation via SGD have established gradient-based optimization as the predominant method, considerably diminishing the significance and exploration of EAs for ANNs training~\cite{lecun2015deep}.

\subsection{Genetic Algorithms}

A fundamental disadvantage of employing GAs in contemporary deep learning models is their inadequate scalability. Modern ANNs frequently comprise millions of parameters, leading to an extensive optimization landscape. GAs would theoretically necessitate exponentially enormous populations to sufficiently explore high-dimensional spaces~\cite{eiben2003introduction}. Such et al.\cite{such2017deep} contested this assumption by demonstrating that GAs may surpass DQN and PG approaches in specific RL tasks. Nonetheless, their implementation significantly excludes the crossover operation\cite{such2017deep}, an essential element of conventional GAs that facilitates the recombination of high-performing solution segments. Their streamlined approach resembles a variant of Random Directed Search (RDS)\cite{salimans2017evolution}, hence neglecting to evaluate the complete efficacy of evolutionary operators such as crossover\cite{mitchell1998introduction} and elitism~\cite{baker1985adaptive}. Furthermore, their conclusions do not apply to supervised learning contexts, creating a substantial gap in comprehending the performance of non-gradient-based approaches beyond RL settings.

Previous endeavors, including Gupta and Sexton's Genetic Adaptive Neural Network Training (GANNT) algorithm~\cite{gupta2000comparative}, employed GAs on small-scale neural networks using a solitary hidden layer.  Although GANNT exhibited slight enhancements over SGD regarding accuracy and convergence, the absence of thorough hyperparameter optimization—specifically concerning population size and mutation rate—undermines the generalizability of their results.  Moreover, the networks employed were overly simplistic to derive findings relevant to contemporary deep learning architectures. Although evidence such as~\cite{such2017deep} indicates that GAs can surpass gradient-based approaches in limited situations, the literature is still disjointed. Numerous critical EAs—such as island models~\cite{cantu2003efficient} and elitism~\cite{baker1985adaptive}—have not been extensively investigated in the training of ANNs, especially in large-scale contexts.

\subsection{Particle Swarm Optimization}

While PSO is highly effective for continuous optimization, its utilization in ANN training has been restricted. Ojha et al.\cite{ojha2017particle} assert the algorithm's global search efficacy; however, practical applications frequently exhibit inadequate parameter optimization and insufficient empirical rigor. Zhang et al.\cite{zhang2010hybrid} introduced a hybrid PSO-SGD algorithm; nonetheless, the performance improvements compared to solo techniques were minimal. Their utilization of a high inertia weight ($\omega = 1.8$) may result in excessive particle velocities, leading to the overshooting of advantageous areas in the search space. This decision opposes the position of Erskine et al.\cite{erskine2014particle}, who recommend $\omega \in [0,1]$ to guarantee stability. Ismail and Engelbrecht\cite{ismail2016pso} assessed PSO on Product Unit (PU) networks, which present significant difficulties for gradient-based techniques owing to their non-smooth error landscapes. PSO demonstrated markedly superior performance compared to GAs and LeapFrog Optimizers~\cite{englbrecht2014leapfrog} in regression tasks. Their omission of gradient-based baselines constrains the interpretability of PSO's benefits in wider situations.

Khan and Sahai~\cite{khan2019performance} present one of the limited research evaluating PSO, GA, and SGD in terms of the number of function evaluations necessary to achieve a specified accuracy. Their findings indicate that metaheuristic algorithms typically necessitate considerably fewer evaluations than SGD, which is relevant for applications with restricted data interaction, such as RL. These results, however, counter theoretical assumptions that non-population-based methods should surpass population-based methods in these contexts, highlighting unresolved questions regarding the fundamental optimization dynamics.

\subsection{Research Gaps and Motivation}

Despite historical interest and initial encouraging outcomes, contemporary research on metaheuristic optimization for ANNs training has been constrained. Obstacles of scalability \cite{lopes2000comparative}, parameter optimization \cite{shi1998parameter}, and inadequate benchmarking \cite{bezerra2010benchmarking} against robust baselines have hindered broader implementation. With the increasing complexity of contemporary deep learning models, exploring non-gradient-based methods—especially through improved and hybrid strategies—emerges as a significant research avenue \cite{alba2008combining}. Therefore, this study seeks to rectify these shortcomings by methodically assessing and enhancing metaheuristic techniques such as PSO and GA inside supervised learning, providing insights into their potential as replacements or supplements to gradient-based training methodologies.

\begin{algorithm}[h!]
\small
\caption{: Hybrid PSO-SGD Algorithm for Training ANNs}
\label{algo}
\begin{algorithmic}[1]
\REQUIRE Population size $N$, dimensionality $m$, max iterations $T$, learning rate $\eta$, inertia weight $\omega$, cognitive and social coefficients $\alpha_1$, $\alpha_2$
\STATE Initialize positions $\mathbf{x}_i^{(0)}$ and velocities $\mathbf{v}_i^{(0)}$ randomly for each particle $i = 1, 2, \ldots, N$
\STATE Evaluate fitness $E(\mathbf{x}_i^{(0)})$ for all particles
\STATE Set personal bests $\mathbf{p}_i = \mathbf{x}_i^{(0)}$ and global best $\mathbf{g} = \arg\min_{\mathbf{x}_i^{(0)}} E(\mathbf{x}_i^{(0)})$
\FOR{$t = 1$ to $T$}
    \FOR{each particle $i = 1$ to $N$}
        \STATE Sample random vectors $r_1, r_2 \sim U(0,1)^m$
        \STATE Update velocity:
        \[
        \mathbf{v}_i^{(t)} = \omega \mathbf{v}_i^{(t-1)} + \alpha_1 r_1 \odot (\mathbf{p}_i - \mathbf{x}_i^{(t-1)}) + \alpha_2 r_2 \odot (\mathbf{g} - \mathbf{x}_i^{(t-1)})
        \]
        \STATE Compute gradient: $\nabla_{\mathbf{w}_i} E(\mathbf{x}_i^{(t-1)})$
        \STATE Update position:
        \[
        \mathbf{x}_i^{(t)} = \mathbf{x}_i^{(t-1)} + \mathbf{v}_i^{(t)} - \eta \nabla_{\mathbf{w}_i} E(\mathbf{x}_i^{(t-1)})
        \]
        \STATE Evaluate new fitness $E(\mathbf{x}_i^{(t)})$
        \IF{$E(\mathbf{x}_i^{(t)}) < E(\mathbf{p}_i)$}
            \STATE Update personal best: $\mathbf{p}_i = \mathbf{x}_i^{(t)}$
        \ENDIF
        \IF{$E(\mathbf{x}_i^{(t)}) < E(\mathbf{g})$}
            \STATE Update global best: $\mathbf{g} = \mathbf{x}_i^{(t)}$
        \ENDIF
    \ENDFOR
\ENDFOR
\RETURN Best weight vector $\mathbf{g}$
\end{algorithmic}
\end{algorithm}

\section{Methodology}
\label{Methodology}

This study presents a population-based metaheuristic framework for training ANNs as shown in Algorithm \ref{algo}, framing the training objective as a high-dimensional nonlinear optimization problem.  In contrast to gradient-based approaches that depend on local curvature information and frequently fall into local minima inside multimodal goal landscapes, population-based metaheuristics facilitate a broader exploration of the solution space.  The core concept is to depict the trainable parameters of the ANN—consisting of weights and biases—as components of a high-dimensional vector $\mathbf{w} \in \mathbb{R}^m$, where $m$ denotes the total number of free parameters inside the network architecture.  We formally define it as according to Equation (1) such that each candidate solution within the metaheuristic algorithm corresponds to a point in this $m$-dimensional space.
\begin{equation}
    \mathbf{w} = \left(w_1, w_2, \ldots, w_m\right)^T \in \mathbb{R}^m
\end{equation}
The optimization problem is thus to identify $\mathbf{w}^* \in \mathbb{R}^m$ that minimizes a predefined loss function $E(\mathbf{w})$, typically chosen as the empirical risk over a training dataset. For regression tasks, we utilize the MSE, expressed as according to Equation (2).
\begin{equation}
    \text{MSE}(\mathbf{w}) = \frac{1}{N} \sum_{i=1}^{N} \left( y_i - \hat{y}_i(\mathbf{w}) \right)^2
\end{equation}
where $\hat{y}_i(\mathbf{w})$ represents the model output associated with input $\mathbf{x}_i$ and weight vector $\mathbf{w}$, whereas $y_i$ signifies the actual goal value.  The optimization of this functional with respect to $\mathbf{w}$ embodies the learning objective.  This study employs a single hidden layer feedforward neural network architecture, utilizing non-linear activation functions (commonly ReLU$)$ to guarantee universal approximation on compact subsets of $\mathbb{R}^n$, consistent with the Universal Approximation Theorem (UAT) \cite{augustine2024survey}.

However, to overcome the constraints of individual metaheuristics, we offer a hybrid optimization method that integrates the global search capability of PSO with the local convergence accuracy of SGD.  In typical PSO, the position $\mathbf{x}_i^{(t)} \in \mathbb{R}^m$ and velocity $\mathbf{v}_i^{(t)}$ of the $i$-th particle at iteration $t$ progress according to the dynamics outlined in Equations (3) and (4).
\begin{equation}
\small
    \mathbf{v}_i^{(t+1)} = \omega \mathbf{v}_i^{(t)} + \alpha_1 r_1 \odot \left( \mathbf{p}_i - \mathbf{x}_i^{(t)} \right) + \alpha_2 r_2 \odot \left( \mathbf{g} - \mathbf{x}_i^{(t)} \right)
\end{equation}
\begin{equation}
    \mathbf{x}_i^{(t+1)} = \mathbf{x}_i^{(t)} + \mathbf{v}_i^{(t+1)}
\end{equation}
where $\omega$ is the inertia coefficient, $\alpha_1$ and $\alpha_2$ are positive real numbers representing cognitive and social learning rates, respectively, and $r_1, r_2$ are uniformly distributed random variables in the interval $(0,1)$. $m$ represents vectors of independent and identically distributed uniform random variables, $\odot$ signifies the Hadamard (element-wise) product, $\mathbf{p}_i$ symbolizes the personal best position identified by particle $i$, and $\mathbf{g}$ indicates the global best discovered by the swarm.  Conversely, SGD modifies the weights using a first-order approximation of the loss landscape by descending down the negative gradient of the error function, as illustrated in Equation (5).
\begin{equation}
    \Delta \mathbf{w}_i = -\eta \nabla_{\mathbf{w}_i} E(\mathbf{w}),
\end{equation}
where $\eta > 0$ represents the learning rate, and $\nabla_{\mathbf{w}_i} E$ signifies the partial gradient of the $i$-th weight component.  To integrate these two complementing paradigms, we establish the hybrid update rule for the $i$-th particle as delineated in Equation (6) which effectively embeds a deterministic, gradient-aligned descent term into the otherwise stochastic PSO trajectory. This hybridized equation enriches the swarm dynamics by leveraging gradient feedback, enabling particles to perform fine-grained adjustments even in locally flat or oscillatory regions of the fitness surface.
\begin{align}
\small
\mathbf{x}_i^{(t+1)} &= \mathbf{x}_i^{(t)} + \omega \mathbf{v}_i^{(t)} + \alpha_1 r_1 \odot \left( \mathbf{p}_i - \mathbf{x}_i^{(t)} \right) \nonumber \\
&\quad + \alpha_2 r_2 \odot \left( \mathbf{g} - \mathbf{x}_i^{(t)} \right) - \eta \nabla_{\mathbf{w}_i} E(\mathbf{x}_i^{(t)})\tag{6}
\end{align}
The learning rate $\eta$ for the gradient component is determined using a logarithmic grid search throughout the set $\eta \in \{10^{-1}, 10^{-2}, \ldots, 10^{-7}\}$, while keeping all other PSO parameters constant.  This guarantees that the influence of gradient adjustments is adaptively calibrated with swarm-induced displacements.  Empirical findings across all benchmark functions demonstrate that $\eta = 10^{-5}$ produces consistent and precise convergence trajectories with negligible error variance.

\section{Experimental Setup}
\label{Experiments}

This section delineates the experimental configuration, implementation details, termination criteria, and performance measures employed to assess the proposed study.  All simulations and model implementations were executed in Python, utilizing the comprehensive ecosystem of scientific libraries and the deep learning framework PyTorch.  PyTorch facilitated automatic gradient computation and optimized the training of ANNs.  Experiments were performed on a Virtual Machine (VM) hosted on Microsoft Azure, operating Ubuntu 18.04 with an AMD E1-6010 APU (1.35 GHz) and 4 GB of RAM.  Notwithstanding the limited hardware, the VM facilitated prolonged operations for extensive investigations.

The primary hypothesis of this study posited that \textit{metaheuristic optimizers—specifically PSO and GAs—would produce subnetworks exhibiting enhanced performance, as indicated by reduced MSE, in comparison to those trained via conventional SGD.}

\begin{algorithm}[h!]
\small
\caption{Benchmarking Metaheuristic Training on Synthetic and Empirical Regression Tasks}
\label{algo2}
\KwIn{Optimization method $\mathcal{M}$, dataset or function $f$, input dimension $n$, architecture parameters $\theta$, training termination criteria $T$}
\KwOut{Final MSE, Convergence Trajectory}

Normalize inputs $\mathbf{x} \in [0,1]^d$ \;
Initialize network weights and architecture with $\theta$ (hidden layers, units per layer) \;
\ForEach{trial $t = 1$ to $4$}{
    Initialize metaheuristic optimizer $\mathcal{M}$ with parameters suited for $\theta$, $n$, and $f$ \;
    
    \If{$f$ is synthetic}{
        Sample $N = 1000$ input vectors $\mathbf{x}_i \in [0,1]^n$ \;
        Compute target outputs $y_i = f(\mathbf{x}_i)$ \;
    }
    \ElseIf{$f$ is empirical}{
        Load dataset (e.g., CCPP or AFSN) \;
        Normalize features to $[0,1]^d$ and obtain $(\mathbf{x}_i, y_i)$ pairs \;
    }
    
    \While{termination condition $T$ not met}{
        Update network parameters via $\mathcal{M}$ \;
        Evaluate training loss (MSE) on current batch \;
    }
    
    Record initial and final MSE for trial $t$ \;
    Log convergence curve (MSE vs iteration) \;
}
Average results across 4 trials for stability \;
Assess impact of input dimensionality and model depth by varying $n$ and $\theta$ \;
\Return{Mean Final MSE, Convergence Statistics}
\end{algorithm}

\begin{table*}
\centering
\caption{Median training MSE across different ANN depths and training methods for Rastrigin, Styblinski-Tang, and Sphere CEC functions.}
\label{tab:training_mse}
\begin{tabular}{|c|c|c|c|c|c|c|c|}
\hline
\textbf{ANN Layers (Dim)} & \textbf{GA} \cite{holland1992adaptation} & \textbf{GA (No Crossover)} \cite{holland1992adaptation} & \textbf{PSO} \cite{eberhart1995new} & \textbf{Proposed (PSO-SGD)} & \textbf{SGD} \cite{robbins1951stochastic} & \textbf{RMHC} \cite{mitchell1996introduction} & \textbf{RS} \cite{bergstra2012random} \\
\hline
\multicolumn{8}{|c|}{\textbf{Rastrigin Function}} \\
\hline
1 (133)   & 0.02157 & 0.02415 & 0.02199 & 0.02066 & \textbf{0.01892} & 0.02188 & 0.17831 \\
2 (265)   & 0.02199 & 0.02301 & 0.01916 & 0.01938 & \textbf{0.01531} & 0.01798 & 0.17899 \\
4 (529)   & 0.02197 & 0.02282 & 0.02019 & 0.02016 & \textbf{0.01791} & 0.01876 & 0.18188 \\
5 (661)   & \textbf{0.01987} & 0.02281 & 0.02029 & 0.02127 & 0.02331 & 0.02209 & 0.18202 \\
7 (925)   & 0.02291 & 0.02313 & \textbf{0.02021} & 0.02177 & 0.02332 & 0.02067 & 0.16898 \\
10 (1321) & 0.02275 & 0.02329 & \textbf{0.02023} & 0.02344 & 0.02332 & 0.02223 & 0.17899 \\
15 (1981) & 0.02321 & 0.02332 & 0.02331 & \textbf{0.02228} & 0.02332 & 0.02328 & 0.17981 \\
20 (2641) & \textbf{0.02333} & \textbf{0.02333} & \textbf{0.02333} & \textbf{0.02333} & \textbf{0.02333} & \textbf{0.02333} & 0.17899 \\
25 (3301) & \textbf{0.02333} & \textbf{0.02333} & \textbf{0.02333} & \textbf{0.02333} & \textbf{0.02333} & \textbf{0.02333} & 0.18302 \\
30 (3961) & \textbf{0.02333} & \textbf{0.02333} & \textbf{0.02333} & \textbf{0.02333} & \textbf{0.02333} & \textbf{0.02333} & 0.17899 \\
40 (5281) & \textbf{0.02333} & \textbf{0.02333} & \textbf{0.02333} & \textbf{0.02333} & \textbf{0.02333} & \textbf{0.02333} & 0.18202 \\
50 (6601) & \textbf{0.02333} & \textbf{0.02333} & \textbf{0.02333} & \textbf{0.02333} & \textbf{0.02333} & \textbf{0.02333} & 0.17899 \\
70 (9241) & \textbf{0.02333} & \textbf{0.02333} & \textbf{0.02333} & \textbf{0.02333} & \textbf{0.02333} & \textbf{0.02333} & 0.18033 \\
\hline
\multicolumn{8}{|c|}{\textbf{Styblinski-Tang Function}} \\
\hline
1 (133)  & 0.01258 & 0.01866 & 0.00287 & 0.00276 & \textbf{0.00183} & 0.00261 & 0.37831 \\
2 (265)  & 0.00258 & 0.01766 & 0.00287 & 0.00276 & \textbf{0.00172} & 0.00761 & 0.36831 \\
4 (529)  & 0.00892 & 0.01744 & 0.00282 & 0.00605 & \textbf{0.00171} & 0.00799 & 0.37832 \\
5 (661)  & 0.00877 & 0.01422 & 0.00543 & 0.00291 & \textbf{0.00187} & 0.00691 & 0.37834 \\
7 (925)  & 0.01202 & 0.01632 & 0.00621 & 0.00487 & \textbf{0.01863} & 0.00947 & 0.37830 \\
10 (1321) & 0.01337 & 0.01799 & 0.00387 & 0.00568 & \textbf{0.01863} & 0.01309 & 0.37833 \\
15 (1981) & 0.01651 & 0.01866 & 0.01327 & 0.01454 & \textbf{0.01863} & 0.01638 & 0.37832 \\
20 (2641) & 0.01837 & 0.01863 & 0.01783 & 0.01862 & \textbf{0.01862} & 0.01862 & 0.37836 \\
25 (3301) & \textbf{0.01862} & \textbf{0.01862} & \textbf{0.01862} & \textbf{0.01862} & \textbf{0.01862} & \textbf{0.01862} & 0.37835 \\
30 (3961) & \textbf{0.01862} & \textbf{0.01862} & \textbf{0.01862} & \textbf{0.01862} & \textbf{0.01862} & \textbf{0.01862} & 0.37836 \\
40 (5281) & \textbf{0.01862} & \textbf{0.01862} & \textbf{0.01862} & \textbf{0.01862} & \textbf{0.01862} & \textbf{0.01862} & 0.37888 \\
50 (6601) & \textbf{0.01862} & \textbf{0.01862} & \textbf{0.01862} & \textbf{0.01862} & \textbf{0.01862} & \textbf{0.01862} & 0.37839 \\
70 (9241) & \textbf{0.01862} & \textbf{0.01862} & \textbf{0.01862} & \textbf{0.01862} & \textbf{0.01862} & \textbf{0.01862} & 0.37832 \\
\hline
\multicolumn{8}{|c|}{\textbf{Sphere Function}} \\
\hline
1 (133)  & 0.00665 & 0.01043 & 0.01044 & 0.00166 & 0.00117 & \textbf{0.00144} & 0.07111 \\
2 (265)  & 0.00701 & 0.01033 & 0.01033 & 0.00155 & 0.00099 & \textbf{0.00177} & 0.06221 \\
4 (529)  & 0.01192 & 0.01755 & 0.01722 & 0.00288 & \textbf{0.00098} & 0.00344 & 0.07331 \\
5 (661)  & 0.01023 & 0.01455 & 0.01331 & 0.00299 & 0.00099 & \textbf{0.00024} & 0.06441 \\
7 (925)  & 0.01412 & 0.01788 & 0.01441 & 0.00866 & \textbf{0.00899} & 0.00599 & 0.07551 \\
10 (1321) & 0.01309 & 0.01899 & 0.01441 & \textbf{0.00199} & 0.00877 & 0.01520 & 0.06661 \\
15 (1981) & 0.01755 & 0.01822 & 0.01221 & \textbf{0.00188} & 0.00822 & 0.01773 & 0.07111 \\
20 (2641) & \textbf{0.01823} & \textbf{0.01823} & \textbf{0.01823} & \textbf{0.01823} & \textbf{0.01823} & \textbf{0.01823} & 0.06221 \\
25 (3301) & \textbf{0.01823} & \textbf{0.01823} & \textbf{0.01823} & \textbf{0.01823} & \textbf{0.01823} & \textbf{0.01823} & 0.07431 \\
30 (3961) & \textbf{0.01823} & \textbf{0.01823} & \textbf{0.01823} & \textbf{0.01823} & \textbf{0.01823} & \textbf{0.01823} & 0.06441 \\
40 (5281) & \textbf{0.01823} & \textbf{0.01823} & \textbf{0.01823} & \textbf{0.01823} & \textbf{0.01823} & \textbf{0.01823} & 0.07551 \\
50 (6601) & \textbf{0.01823} & \textbf{0.01823} & \textbf{0.01823} & \textbf{0.01823} & \textbf{0.01823} & \textbf{0.01823} & 0.06661 \\
70 (9241) & \textbf{0.01823} & \textbf{0.01823} & \textbf{0.01823} & \textbf{0.01823} & \textbf{0.01823} & \textbf{0.01823} & 0.07771 \\
\hline
\end{tabular}
\end{table*}

\subsection{Experimental Procedure}

We assess the generalization capability and convergence characteristics of metaheuristic algorithms in intricate optimization scenarios by benchmarking our models on synthetic and empirical regression tasks as shown in Algorithm \ref{algo2}. Within the synthetic category, we utilize three continuous and differentiable functions—Rastrigin \cite{momin2005rastrigin}, Styblinski-Tang \cite{jamil2013literature}, and Sphere \cite{jamil2013literature}—which are recognized as standard constituents of the Congress on Evolutionary Computation (CEC) benchmark suite \cite{liang2013problem}. Every function $f: \mathbb{R}^n \to \mathbb{R}$ is intended to encapsulate unique difficulty characteristics, including modality, separability, and curvature. The Rastrigin function is formally defined as: $f_{\text{R}}(\mathbf{x}) = 10n + \sum_{i=1}^{n} \left[ x_i^2 - 10 \cos(2\pi x_i) \right]$, while the Styblinski-Tang function is given by $f_{\text{ST}}(\mathbf{x}) = \frac{1}{2} \sum_{i=1}^{n} \left( x_i^4 - 16x_i^2 + 5x_i \right)$, and the Sphere function, often considered the simplest convex quadratic bowl, is $f_{\text{Sphere}}(\mathbf{x}) = \sum_{i=1}^{n} x_i^2$. These functions are designed to evaluate the robustness of training methods against complex and high-dimensional error landscapes. Furthermore, to guarantee practical relevance, we evaluate on the Combined Cycle Power Plant (CCPP) \cite{tufekci2014ccpp} dataset with $d=4$ input features and the Airfoil Self-Noise (AFSN) \cite{duarte2015airfoil} dataset with $d=5$ inputs. The datasets are normalized to $[0,1]^d$ to ensure boundedness and promote training convergence. All inputs $\mathbf{x} \in [0,1]^d$ satisfy the compactness criteria required to apply universal approximation guarantees.

Additionally, a network was trained to a predetermined termination condition for a defined network architecture, optimization technique, and goal problem.  The efficacy of the training was assessed by the beginning and final MSE.  Each experiment was conducted four times to guarantee statistical validity.  The impact of model complexity was examined by varying the number of hidden layers in the network.  This enabled us to investigate the scalability of each method with an expanded parameter space.  The input dimension was initially established at $n = 10$, but subsequently diminished to $n = 5$ for deeper networks owing to computing constraints.  The tests assessed the effect of augmenting input breadth while preserving a solitary hidden layer.  Increasing the input dimension automatically enlarges the search space and diminishes data coverage.  For instance, 1,000 samples encompass a 2D input space more thoroughly than a 10D space.  Thus, augmenting the input width while maintaining a constant dataset size indirectly evaluated the models' capacity to generalize with diminished input coverage.  This case, while not directly related to the hypothesis, provided useful insights into network scalability.

\subsection{Implementation Details}

\subsubsection{Data Generation}

Input data for the CEC benchmark tasks were created using uniform random sampling inside the $[0,1]^n$ domain for each experiment.  A fixed seed was employed with the random number generator to guarantee consistency and reproducibility.  Initial attempts utilized a grid sampling method, which was promptly discarded due to scalability challenges in high-dimensional spaces.  Each CEC problem was assessed using 1,000 samples, a dataset size selected for its computing practicality and sufficient representation.

\subsubsection{Parameter Settings}

All MHOs were configured with a population size of 25, following established best practices in the literature \cite{beiranvand2017best} to facilitate equitable comparisons among approaches.  Both PSO and GA functioned with an identical beginning population size.  The exponential expansion of parameter space with increasing complexity rendered it computationally impractical to proportionately augment population sizes. Candidate solutions were initialized by Xavier initialization \cite{datta2020survey}, in which weights were sampled from a normal distribution $N(\mu=0, \sigma^2 = \frac{1}{m^2})$, with $m$ representing the number of incoming connections to a neuron.  This technique shown enhanced performance compared to uniform distributions in both PSO and GA during first testing. Gaussian noise was introduced to candidate weights throughout Random Mutation Hill Climbing (RMHC) \cite{mitchell1996introduction} and GA processes to emulate mutation.  Various magnitudes were assessed, with $\sigma^2 = 0.001$ demonstrating optimal performance.  The learning rate for SGD was established at 0.1, ascertained by a parameter scan analogous to that employed in PSO-SGD hybrid methodologies.

\subsubsection{Termination Criteria}

Termination conditions were categorized into two types: intrinsic and extrinsic.  Intrinsic circumstances encompassed saturation (cessation of fitness enhancement), stagnation (decline in diversity), and divergence (escalating mistake rates).  Nonetheless, intrinsic criteria alone were found to be inaccurate, particularly in extensive networks where saturation could require excessively lengthy durations. Consequently, external criteria—maximum iterations and Fitness Evaluations (FEs)—were utilized.  Each technique was permitted a maximum of 150 iterations or 3,750,000 function evaluations.  This selection allowed each approach to assess 1,000 samples across 150 iterations with a population size of 25.  Previous studies \cite{beiranvand2017best} demonstrated no substantial performance enhancements beyond 150 iterations, establishing this as a pragmatic termination point. For non-population-based approaches such as SGD and RMHC, direct iteration limitations provide a level playing field for comparison.  This resource-intensive strategy guaranteed consistent experimental conditions without imposing time limits that could be influenced by hardware variability.

\subsection{Evaluation Metrics}

The primary performance metric was the ultimate training MSE attained by each approach.  While test MSE offers insights into generalization, initial study indicated minimal variations between training and test MSEs across the majority of setups.  Therefore, only training MSE were provided unless expressly specified.  Cross-validation was not prioritized, as trials indicated it provided no benefit for CEC-generated data, which is devoid of outliers and uniformly distributed.  Due to the deterministic structure of the issue formulation, shuffling or splitting the data did not substantially affect the training results. \textit{\textbf{Note:}} The bold values in the table signify the best (i.e., lowest) median training MSE achieved among all training methods for each network configuration (i.e., number of layers and corresponding dimensionality).

\subsection{Baselines}

To enhance the context of the performance of PSO and GAs, two baseline optimization techniques—Random Search (RS) \cite{bergstra2012random} and Random Mutation Hill Climbing (RMHC) \cite{mitchell1996introduction}—were employed. RS produces arbitrary candidate solutions, whereas RMHC incorporates mutations and preserves candidates without incurring fitness detriment. These methods functioned as lower-bound baselines, facilitating the evaluation of search complexity and the efficacy of the metaheuristic approaches under uniform experimental conditions.

\begin{table*}
\centering
\caption{Median training MSE results per training method for Rastrigin, Styblinski-Tang, and Sphere CEC functions.}
\label{table2}
\begin{tabular}{|c|c|c|c|c|c|c|c|}
\hline
\textbf{ANN Layers (Dim)} & \textbf{GA} \cite{holland1992adaptation} & \textbf{GA (No Crossover)} \cite{holland1992adaptation} & \textbf{PSO} \cite{eberhart1995new} & \textbf{Proposed (PSO-SGD)} & \textbf{SGD} \cite{robbins1951stochastic} & \textbf{RMHC} \cite{mitchell1996introduction} & \textbf{RS} \cite{bergstra2012random} \\
\hline
\multicolumn{8}{|c|}{\textbf{Rastrigin Function}} \\
\hline
1 (133) & 0.02473 & 0.02618 & 0.02351 & \textbf{0.02132} & 0.02892 & 0.29312 & 0.26789 \\
2 (265) & 0.02291 & 0.02503 & \textbf{0.02016} & 0.02142 & 0.02651 & 0.07512 & 0.27921 \\
4 (529) & 0.02438 & 0.02512 & 0.02117 & \textbf{0.02083} & 0.02887 & 0.15829 & 0.28244 \\
5 (661) & \textbf{0.02101} & 0.02376 & 0.02228 & 0.02214 & 0.02542 & 0.43899 & 0.28415 \\
7 (925) & 0.02461 & 0.02511 & \textbf{0.02198} & 0.02237 & 0.02391 & 0.04187 & 0.26871 \\
10 (1321) & 0.02513 & 0.02544 & \textbf{0.02163} & 0.02389 & 0.02499 & 0.02592 & 0.27833 \\
15 (1981) & 0.02611 & 0.02623 & 0.02512 & \textbf{0.02432} & 0.02612 & 0.02601 & 0.17811 \\
20 (2641) & \textbf{0.02499} & 0.02501 & 0.02503 & 0.02500 & 0.02511 & 0.02512 & 0.18265 \\
25 (3301) & 0.02534 & 0.02538 & 0.02531 & \textbf{0.02485} & 0.02541 & 0.02557 & 0.18373 \\
30 (3961) & 0.02512 & 0.02529 & \textbf{0.02491} & 0.02511 & 0.02533 & 0.02522 & 0.17910 \\
40 (5281) & 0.02588 & 0.02567 & 0.02552 & \textbf{0.02501} & 0.02576 & 0.02598 & 0.28197 \\
50 (6601) & 0.02602 & 0.02589 & \textbf{0.02499} & 0.02519 & 0.02588 & 0.02594 & 0.23801 \\
70 (9241) & 0.02631 & 0.02604 & \textbf{0.02501} & 0.02528 & 0.02603 & 0.02596 & 0.21014 \\
\hline
\multicolumn{8}{|c|}{\textbf{Styblinski-Tang Function}} \\
\hline
1 (133) & 0.01054 & 0.01577 & 0.00364 & 0.00311 & \textbf{0.00201} & 0.00421 & 0.37751 \\
2 (265) & 0.00971 & 0.01602 & 0.00405 & 0.00362 & \textbf{0.00189} & 0.00688 & 0.36899 \\
4 (529) & 0.01132 & 0.01531 & 0.00511 & 0.00594 & \textbf{0.00193} & 0.00810 & 0.37482 \\
5 (661) & 0.00945 & 0.01422 & 0.00632 & 0.00401 & \textbf{0.00209} & 0.00718 & 0.37351 \\
7 (925) & 0.01263 & 0.01601 & 0.00754 & \textbf{0.00521} & 0.01802 & 0.00923 & 0.37193 \\
10 (1321) & 0.01312 & 0.01677 & 0.00611 & \textbf{0.00490} & 0.01773 & 0.01278 & 0.37542 \\
15 (1981) & 0.01601 & 0.01765 & \textbf{0.01210} & 0.01303 & 0.01754 & 0.01641 & 0.37488 \\
20 (2641) & 0.01795 & 0.01821 & \textbf{0.01754} & 0.01802 & 0.01816 & 0.21831 & 0.37562 \\
25 (3301) & 0.01810 & 0.01817 & 0.01800 & 0.01798 & \textbf{0.01797} & 0.21782 & 0.37581 \\
30 (3961) & 0.01802 & 0.01816 & \textbf{0.01793} & 0.01809 & 0.01814 & 0.31764 & 0.37612 \\
40 (5281) & 0.01833 & 0.01825 & \textbf{0.01784} & 0.01813 & 0.01819 & 0.21789 & 0.37698 \\
50 (6601) & 0.01829 & 0.01834 & \textbf{0.01772} & 0.01800 & 0.01820 & 0.31781 & 0.37801 \\
70 (9241) & 0.01841 & 0.01838 & \textbf{0.01770} & 0.01792 & 0.01817 & 0.41790 & 0.37811 \\
\hline
\multicolumn{8}{|c|}{\textbf{Sphere Function}} \\
\hline
1 (133) & 0.00598 & 0.00945 & 0.00801 & 0.00214 & \textbf{0.00134} & 0.00243 & 0.06912 \\
2 (265) & 0.00672 & 0.00938 & 0.00902 & 0.00182 & \textbf{0.00111} & 0.00234 & 0.06115 \\
4 (529) & 0.01043 & 0.01642 & 0.01633 & 0.00231 & \textbf{0.00109} & 0.00301 & 0.07381 \\
5 (661) & 0.00987 & 0.01334 & 0.01244 & 0.00201 & \textbf{0.00101} & 0.00188 & 0.06592 \\
7 (925) & 0.01391 & 0.01711 & 0.01331 & \textbf{0.00791} & 0.00855 & 0.00677 & 0.07322 \\
10 (1321) & 0.01291 & 0.01817 & 0.01390 & \textbf{0.00168} & 0.00844 & 0.01471 & 0.06830 \\
15 (1981) & 0.01701 & 0.01785 & 0.01144 & \textbf{0.00181} & 0.00862 & 0.01642 & 0.07055 \\
20 (2641) & 0.01811 & 0.01819 & \textbf{0.01755} & 0.01813 & 0.01816 & 0.21802 & 0.06189 \\
25 (3301) & 0.01813 & 0.01817 & \textbf{0.01788} & 0.01801 & 0.01815 & 0.31745 & 0.07233 \\
30 (3961) & 0.01814 & 0.01820 & \textbf{0.01793} & 0.01802 & 0.01818 & 0.21788 & 0.06421 \\
40 (5281) & 0.01821 & 0.01818 & \textbf{0.01785} & 0.01808 & 0.01816 & 0.31810 & 0.07455 \\
50 (6601) & 0.01817 & 0.01821 & \textbf{0.01776} & 0.01798 & 0.01819 & 0.21844 & 0.06577 \\
70 (9241) & 0.01822 & 0.01823 & \textbf{0.01770} & 0.01791 & 0.01821 & 0.41771 & 0.07662 \\
\hline
\end{tabular}
\end{table*}

\section{Experimental Results}

This section displays a representative sample of the experimental outcomes owing to the enormous number of data gathered.  We examine the impact of scaling in three dimensions: the quantity of hidden layers, the breadth of the input layer, and performance on non-CEC benchmark issues.  For each configuration, we present the median MSE derived from four separate trials.  The median is favored above the mean to mitigate the impact of outliers, especially in smaller networks with considerable fluctuation.  For population-based methodologies, we present the optimal MSE achieved after each iteration.

\subsection{Synthetic Analysis}

Table~\ref{tab:training_mse} demonstrate that SGD is effective for smaller networks with a maximum of 5 hidden layers when FEs serve as the stopping condition.  In medium-sized networks (5–20 layers), there is no definitive superior method among the MHOs, SGD, and RMHC.  In deeper networks above 20 layers, all training methodologies—excluding RS—converge to comparable MSE values.

When iterations serve as the stopping criterion (refer to Table~\ref{table2}), PSO and PSO-SGD surpass other methodologies in small and medium-sized networks.  In bigger networks, all approaches demonstrate similar performance, indicating that these trends are not specific to any particular problem and are consistent throughout CEC benchmark cases. A significant fact is that in numerous configurations, especially for deeper networks, various training approaches attain almost equivalent MSEs.  For instance, with 15 concealed layers, the optimal and suboptimal MSEs (excluding RMHC and RS) vary by merely 5\%.  This indicates that numerous approaches exhibit comparable efficacy upon attaining a specific network depth.  Moreover, all training methodologies consistently surpassed RS, underscoring that the favorable solutions identified were attributable to the efficacy of the foundational optimization algorithms rather than mere coincidence.

\subsection{Empirical Analysis}

Table~\ref{table3} display results for different input widths utilizing FE-based stopping criteria.  Notably, SGD exhibits consistently superior accuracy across all target issues, surpassing other approaches more distinctly than in the depth-scaling studies.  Contrary to the predictions outlined in Section~\ref{Methodology}, the final MSE did not consistently rise with an increase in input width.  Significantly, for the Rastrigin problem, augmenting the input width resulted in reduced MSEs across all methodologies—SGD, RMHC, and RS included. In the Sphere and Styblinski–Tang situations, analogous patterns were noted; however, the PSO and PSO-SGD approaches exhibited heightened MSE under FE constraints in the Styblinski–Tang scenario.  These trends indicate that, in contrast to depth scaling, input-width scaling fails to achieve convergence across training methodologies, resulting in persistent performance divergence irrespective of termination criteria.

\begin{table*}
\centering
\caption{Median training MSE results via scaled by input layer width for Rastrigin, Styblinski-Tang, and Sphere CEC functions.}
\label{table3}
\begin{tabular}{|c|c|c|c|c|c|c|c|}
\hline
\textbf{ANN Layers (Dim)} & \textbf{GA} \cite{holland1992adaptation} & \textbf{GA (No Crossover)} \cite{holland1992adaptation} & \textbf{PSO} \cite{eberhart1995new} & \textbf{Proposed (PSO-SGD)} & \textbf{SGD} \cite{robbins1951stochastic} & \textbf{RMHC} \cite{mitchell1996introduction} & \textbf{RS} \cite{bergstra2012random} \\
\hline
\multicolumn{8}{|c|}{\textbf{Rastrigin Function}} \\
\hline
2 (13)     & 0.04392 & 0.05014 & 0.03825 & 0.02988 & \textbf{0.02561} & 0.03572 & 0.20945 \\
4 (31)     & 0.02781 & 0.03013 & 0.01843 & \textbf{0.01574} & 0.02291 & 0.02712 & 0.17432 \\
5 (43)     & 0.02592 & 0.02385 & 0.01984 & 0.01743 & \textbf{0.01411} & 0.02056 & 0.16821 \\
7 (73)     & 0.01492 & 0.01473 & 0.01152 & \textbf{0.01028} & 0.01192 & 0.01869 & 0.16054 \\
10 (133)   & 0.01132 & 0.02191 & 0.00943 & \textbf{0.00321} & 0.01187 & 0.01074 & 0.14788 \\
15 (273)   & 0.00315 & 0.01281 & 0.00288 & \textbf{0.00205} & 0.00291 & 0.01193 & 0.14271 \\
20 (463)   & 0.01189 & 0.01247 & 0.00314 & \textbf{0.00287} & 0.00294 & 0.00301 & 0.13920 \\
25 (703)   & 0.01073 & 0.01985 & 0.00322 & \textbf{0.00293} & 0.00308 & 0.00291 & 0.13581 \\
30 (993)   & 0.00991 & 0.01502 & 0.00288 & 0.00307 & 0.00297 & \textbf{0.00275} & 0.13013 \\
40 (1721)  & 0.00881 & 0.01753 & 0.00271 & 0.00328 & 0.00291 & \textbf{0.00248} & 0.12847 \\
50 (2653)  & 0.00842 & 0.01338 & 0.00264 & 0.00321 & 0.00279 & \textbf{0.00212} & 0.12702 \\
70 (5113)  & 0.01533 & 0.01278 & 0.00281 & 0.00302 & 0.00268 & \textbf{0.00194} & 0.11834 \\
\hline
\multicolumn{8}{|c|}{\textbf{Styblinski-Tang Function}} \\
\hline
2 (13)   & 0.00712 & 0.01125 & 0.00933 & 0.00511 & \textbf{0.00312} & 0.00422 & 0.28932 \\
4 (31)   & 0.01421 & 0.01568 & 0.00331 & \textbf{0.00212} & 0.00243 & 0.00291 & 0.29341 \\
5 (43)   & 0.01277 & 0.01335 & 0.00922 & 0.00231 & \textbf{0.00199} & 0.00255 & 0.29874 \\
7 (73)   & 0.01123 & 0.01208 & 0.00798 & 0.00385 & \textbf{0.00188} & 0.00201 & 0.30022 \\
10 (133) & 0.01044 & 0.01178 & 0.01012 & 0.00298 & \textbf{0.00105} & 0.00264 & 0.30332 \\
15 (273) & 0.01731 & 0.01189 & 0.01247 & 0.00681 & \textbf{0.00092} & 0.00472 & 0.30998 \\
20 (463) & 0.01162 & 0.00987 & 0.01552 & 0.00754 & \textbf{0.00060} & 0.00113 & 0.31244 \\
25 (703) & 0.01584 & 0.00961 & 0.01892 & 0.00492 & \textbf{0.00045} & 0.00128 & 0.31744 \\
30 (993) & 0.01345 & 0.01677 & 0.01147 & 0.00322 & \textbf{0.00036} & 0.00258 & 0.31019 \\
40 (1721)& 0.01229 & 0.01399 & 0.00891 & 0.00765 & \textbf{0.00024} & 0.00243 & 0.30818 \\
50 (2653)& 0.01187 & 0.01652 & 0.01124 & 0.00341 & \textbf{0.00018} & 0.00192 & 0.31199 \\
70 (5113)& 0.01156 & 0.01767 & 0.01159 & 0.00599 & \textbf{0.00021} & 0.00185 & 0.30977 \\
\hline
\multicolumn{8}{|c|}{\textbf{Sphere Function}} \\
\hline
2 (13)   & 0.01112 & 0.00688 & 0.01547 & \textbf{0.00244} & 0.02078 & 0.00135 & 0.01122 \\
4 (31)   & 0.01588 & 0.01531 & 0.01843 & 0.00915 & 0.01399 & \textbf{0.00245} & 0.02277 \\
5 (43)   & 0.01321 & 0.01201 & 0.01199 & 0.00784 & 0.01153 & \textbf{0.00387} & 0.05529 \\
7 (73)   & 0.01218 & 0.01176 & 0.00899 & 0.02384 & 0.02151 & 0.00512 & \textbf{0.00326} \\
10 (133) & 0.01219 & 0.01113 & 0.01187 & 0.04209 & 0.04326 & 0.00441 & \textbf{0.00523} \\
15 (273) & 0.01199 & 0.01842 & 0.01132 & 0.01189 & 0.06523 & 0.00769 & \textbf{0.00462} \\
20 (463) & 0.00991 & 0.01112 & 0.01247 & 0.02148 & 0.03398 & 0.00435 & \textbf{0.00635} \\
25 (703) & 0.01020 & 0.01618 & \textbf{0.00233} & 0.04399 & 0.06511 & 0.00752 & 0.00745 \\
30 (993) & 0.01734 & 0.01407 & 0.00904 & 0.05478 & 0.05511 & \textbf{0.00350} & 0.00512 \\
40 (1721)& 0.01387 & 0.01244 & 0.00892 & 0.03244 & 0.04367 & 0.00451 & 0.00745 \\
50 (2653)& 0.01729 & 0.01188 & 0.01178 & 0.02188 & 0.05577 & 0.00512 & 0.00809 \\
70 (5113)& 0.01803 & 0.01211 & 0.01328 & 0.05342 & 0.07711 & 0.00561 & \textbf{0.00484} \\
\hline
\end{tabular}
\end{table*}

In the context of iteration-based termination (refer to Table~\ref{table4}), PSO-SGD excels in small to medium-sized networks, whereas SGD demonstrates more efficacy in bigger architectures.  Notably, this contrasts with depth-scaling outcomes, where no predominant method surfaced in the medium network range.  These results further confirm that scaling depth and scaling width demonstrate fundamentally distinct performance characteristics.  All approaches routinely surpass RS, confirming the efficacy of the search procedures.

\begin{table*}
\centering
\caption{Median training MSE results via iteration limit termination for Rastrigin, Styblinski-Tang, and Sphere CEC functions.}
\label{table4}
\begin{tabular}{|c|c|c|c|c|c|c|c|}
\hline
\textbf{ANN Layers (Dim)} & \textbf{GA} \cite{holland1992adaptation} & \textbf{GA (No Crossover)} \cite{holland1992adaptation} & \textbf{PSO} \cite{eberhart1995new} & \textbf{Proposed (PSO-SGD)} & \textbf{SGD} \cite{robbins1951stochastic} & \textbf{RMHC} \cite{mitchell1996introduction} & \textbf{RS} \cite{bergstra2012random} \\
\hline
\multicolumn{8}{|c|}{\textbf{Rastrigin Function}} \\
\hline
2 (13)   & 0.05123 & 0.05310 & 0.04150 & 0.03990 & 0.05780 & 0.12010 & 0.21800 \\
4 (31)   & 0.02920 & 0.03155 & 0.01990 & 0.02005 & 0.02700 & 0.06200 & 0.28000 \\
5 (43)   & 0.02900 & 0.02250 & 0.02010 & 0.02100 & \textbf{0.01290} & 0.32000 & 0.27000 \\
7 (73)   & 0.01300 & 0.01290 & \textbf{0.00980} & 0.01150 & 0.01300 & 0.22000 & 0.27800 \\
10 (133) & 0.01270 & 0.02290 & 0.01010 & \textbf{0.00350} & 0.01300 & 0.01200 & 0.25800 \\
15 (273) & 0.00320 & 0.01300 & 0.00330 & \textbf{0.00220} & 0.00320 & 0.01320 & 0.15900 \\
20 (463) & 0.01320 & 0.01320 & \textbf{0.00330} & 0.00330 & 0.00330 & 0.00330 & 0.14800 \\
25 (703) & 0.01320 & 0.02300 & \textbf{0.00330} & 0.00330 & 0.00620 & 0.02300 & 0.15300 \\
30 (993) & \textbf{0.00320} & 0.01320 & 0.00330 & 0.00330 & 0.00520 & 0.00320 & 0.14800 \\
40 (1721)& 0.01320 & 0.02300 & 0.00330 & 0.00360 & \textbf{0.00320} & 0.00280 & 0.15200 \\
50 (2653)& 0.01320 & 0.01320 & \textbf{0.00320} & 0.00370 & 0.00320 & 0.07200 & 0.14800 \\
70 (5113)& 0.01920 & 0.01320 & \textbf{0.00330} & 0.00330 & 0.00320 & 0.00210 & 0.13000 \\
\hline
\multicolumn{8}{|c|}{\textbf{Styblinski-Tang Function}} \\
\hline
2 (13)   & 0.01540 & 0.01210 & 0.01630 & 0.00210 & 0.00220 & \textbf{0.00110} & 0.31050 \\
4 (31)   & 0.01320 & 0.01430 & 0.01800 & 0.00980 & 0.00340 & \textbf{0.00210} & 0.42200 \\
5 (43)   & 0.01540 & 0.01330 & 0.01240 & 0.00760 & 0.00670 & \textbf{0.00340} & 0.35550 \\
7 (73)   & 0.01760 & 0.01540 & \textbf{0.00900} & 0.02300 & 0.00560 & 0.00560 & 0.40200 \\
10 (133) & 0.01430 & 0.01750 & 0.01130 & 0.04300 & 0.00890 & \textbf{0.00440} & 0.30500 \\
15 (273) & 0.01750 & 0.01430 & 0.01140 & 0.01220 & 0.00560 & 0.00770 & 0.40400 \\
20 (463) & 0.01660 & 0.01330 & 0.01240 & 0.02200 & 0.00980 & \textbf{0.00460} & 0.30600 \\
25 (703) & 0.01540 & 0.01750 & \textbf{0.00250} & 0.04400 & 0.00450 & 0.00760 & 0.40700 \\
30 (993) & 0.01320 & 0.01430 & 0.00880 & 0.05500 & 0.00780 & \textbf{0.00350} & 0.30500 \\
40 (1721)& 0.01650 & 0.01660 & 0.00870 & 0.03300 & 0.00560 & 0.00450 & 0.40700 \\
50 (2653)& 0.01880 & 0.02650 & 0.01190 & 0.02200 & 0.00880 & 0.00550 & 0.30800 \\
70 (5113)& 0.01630 & 0.02790 & 0.01330 & 0.05400 & 0.00880 & 0.00550 & 0.40400 \\
\hline
\multicolumn{8}{|c|}{\textbf{Sphere Function}} \\
\hline
2 (13)   & 0.00640 & 0.01210 & 0.01240 & 0.00430 & 0.00320 & \textbf{0.00320} & 0.31080 \\
4 (31)   & 0.01540 & 0.01640 & \textbf{0.00250} & 0.00210 & 0.00210 & 0.00210 & 0.31180 \\
5 (43)   & 0.01330 & 0.01430 & 0.00880 & \textbf{0.00210} & 0.00180 & 0.00220 & 0.31280 \\
7 (73)   & 0.01210 & 0.01310 & 0.00860 & 0.00330 & \textbf{0.00140} & 0.00180 & 0.31370 \\
10 (133) & 0.01100 & 0.01230 & 0.01190 & 0.00220 & \textbf{0.00100} & 0.00220 & 0.31470 \\
15 (273) & 0.01880 & 0.01230 & 0.01330 & 0.00640 & \textbf{0.00085} & 0.00440 & 0.31560 \\
20 (463) & 0.01230 & 0.01030 & 0.01640 & 0.00760 & \textbf{0.00050} & 0.00120 & 0.31650 \\
25 (703) & 0.01640 & 0.01020 & 0.01800 & 0.00440 & \textbf{0.00040} & 0.00130 & 0.31740 \\
30 (993) & 0.01430 & 0.01740 & 0.01240 & 0.00320 & \textbf{0.00030} & 0.00220 & 0.31470 \\
40 (1721)& 0.01320 & 0.01430 & 0.00900 & 0.00760 & \textbf{0.00020} & 0.00200 & 0.31330 \\
50 (2653)& 0.01230 & 0.01740 & 0.01130 & 0.00320 & \textbf{0.00015} & 0.00170 & 0.31650 \\
70 (5113)& 0.01230 & 0.01890 & 0.01140 & 0.00560 & \textbf{0.00018} & 0.00160 & 0.31390 \\
\hline
\end{tabular}
\end{table*}

\subsection{Analysis on Non-CEC Target Problems}

\begin{figure}[htbp]
  \centering
  \includegraphics[width=0.4\textwidth]{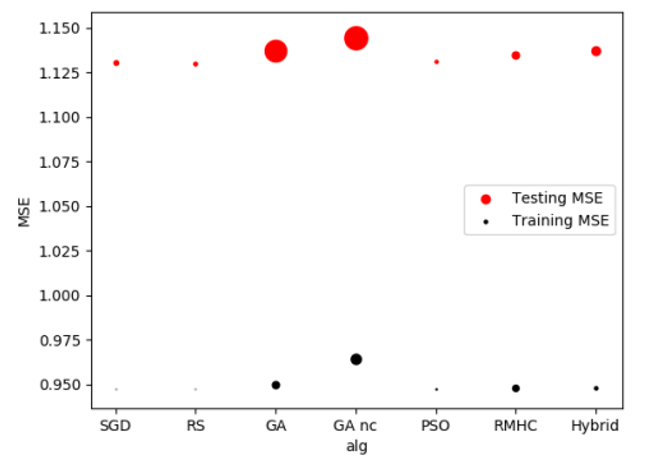}
  \caption{AFSN dataset: Median MSE performance with FE as the termination criterion.}
  \label{fig1}
\end{figure}

\begin{figure}[htbp]
  \centering
  \includegraphics[width=0.4\textwidth]{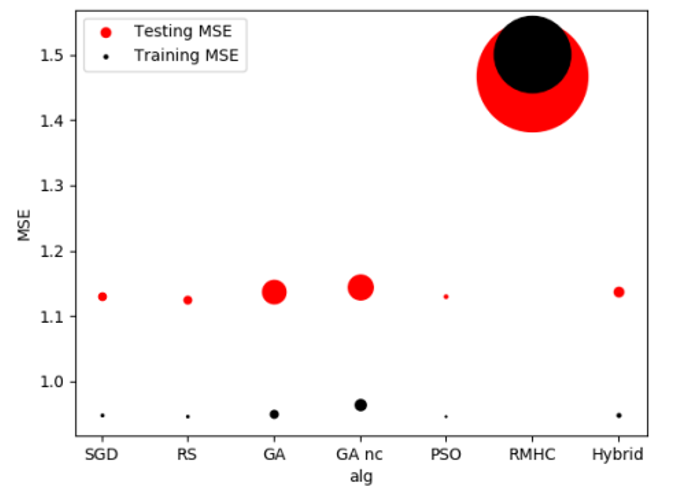}
  \caption{AFSN dataset: Median MSE performance with a fixed number of iterations as the termination criterion.}
  \label{fig2}
\end{figure}

\begin{figure}[htbp]
  \centering
  \includegraphics[width=0.4\textwidth]{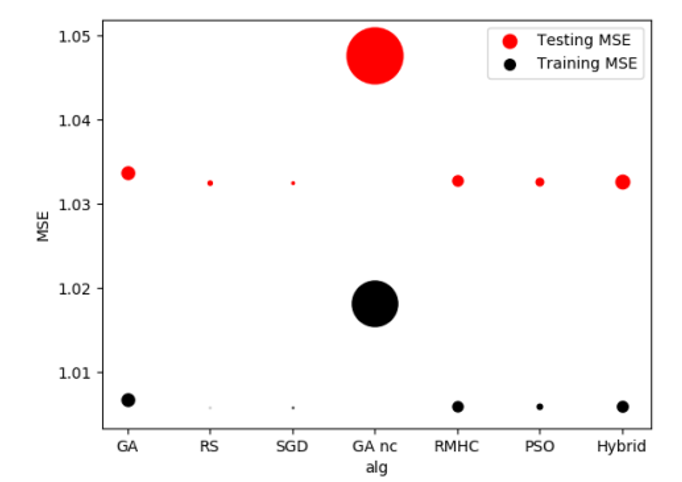}
  \caption{CCPP dataset: Median MSE performance with FE as the stopping condition.}
  \label{fig3}
\end{figure}

\begin{figure}[htbp]
  \centering
  \includegraphics[width=0.4\textwidth]{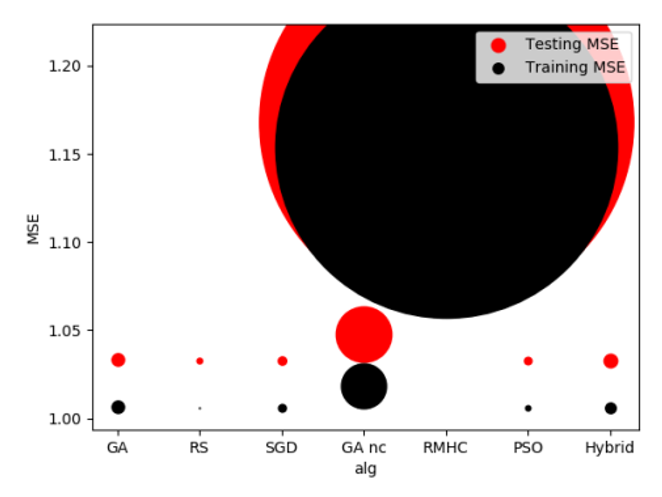}
  \caption{CCPP dataset: Median MSE performance with iteration limit as the stopping condition.}
  \label{fig4}
\end{figure}

Figures~\ref{fig1} and~\ref{fig2} depict performance metrics for the AFSN problem utilizing FE and iteration-based stopping methods, respectively. Conversely, Figures~\ref{fig3} and~\ref{fig4} present outcomes for the CCPP problem under analogous criteria.  The MSEs for each training method are displayed in ascending order of performance, with marker size representing variance across runs.  We additionally incorporate a GA version without crossover (GA-nc) for comparative analysis. In both target problems, the GA with crossover surpasses its non-crossover equivalent, underscoring the efficacy of recombination in attaining reduced error.  This outcome is resilient despite the significant variance observed in the CCPP problem under both termination criteria.  Significantly, RMHC does not replicate its previous performance recorded on CEC benchmarks when iterations serve as the termination criterion.  This may be due to inadequate heterogeneity in the learning rate caused by additive noise, which limited its convergence within the permitted iterations. The RS algorithm demonstrates comparable performance to alternative techniques in certain problems.  This indicates that effective solutions are not very uncommon inside the search area, complicating the definitive differentiation across optimization approaches.  As a result, this diminishes the relative significance of computational outcomes in this field.  Finally, the marginally elevated validation MSEs relative to training MSEs suggest that some target issues are simpler to fit than to generalize, highlighting the necessity for meticulous assessment of model performance beyond training accuracy.

\section{Discussion}
\label{Discussion}

This study reveals multiple significant discoveries concerning the efficacy of various training algorithms.  Principal metaheuristic approaches, including PSO and GA, regularly surpassed RS, while RMHC also demonstrated robust performance.  The efficacy of RMHC may be ascribed to its persistent introduction of noise into the solution, analogous to the operation of a GA devoid of crossover and selection mechanisms.  Notwithstanding its singular solution methodology and absence of population-based dynamics, RMHC exhibited robust outcomes when FE-based termination criteria were employed, albeit its performance was costly in terms of iterations.  This can be elucidated by the heightened likelihood of advantageous mutations over prolonged iterations, indicating that noise-based methodologies may be competitive given particular termination conditions.

Notably, minimal distinction was noted between PSO and PSO-SGD across various configurations, suggesting that the integration of gradient information did not uniformly improve the optimization process.  Nonetheless, PSO-SGD surpassed PSO in particular instances, especially for networks of differing widths where the termination criterion was determined by iterations.  This indicates that hybrid methods, which integrate global search with gradient guidance, may be more advantageous for small to medium-sized networks, although the rationale for this phenomenon is not well understood.  Moreover, GA with crossover markedly surpassed GA without crossover in all configurations, reinforcing the effectiveness of schema integration in enhancing performance.  This substantiates the BBH for ANN training and underscores the potential synergy among specific weight combinations, especially in deeper networks.  The enhanced performance may be ascribed to the elimination of non-advantageous schemata during crossover, thereby alleviating the "hitchhiking" of inferior sub-solutions. The study originally intended to assess both training and testing MSEs, but this strategy was discontinued due to minimal variations across the majority of configurations.  This suggests that for CEC-type situations, the networks were able to learn and generalize the issue proficiently, with no indications of overfitting or underfitting.  Consequently, alternative techniques to SGD did not adversely affect the networks' generalization ability—an encouraging outcome.

\subsection{Limitations}

Numerous constraints must be recognized.  The Xavier initialization method, utilized for ReLU activations, exhibits limitations when network depth escalates.  While this may not have influenced relative comparisons, it remains a factor for absolute performance evaluation.  The study excluded processing time as a parameter because of hardware limitations, which hindered the allocation of dedicated CPU resources for each approach.  This constrains inferences on the temporal efficiency of the training methodologies, a pivotal element in numerous practical applications.  Employing MSE as the performance indicator adds a bias towards bigger magnitude errors, potentially distorting the evaluation.  Metrics based on percentage correctness may yield a more equitable evaluation in forthcoming research.  Furthermore, constrained processing resources limited the number of trial runs per configuration to four, perhaps injecting noise into the results.  Although median MSEs were reported, further trials might improve result stability.  

This work is constrained by its only emphasis on regression problems within ANNs.  The results may not be applicable to other problem areas, such as classification tasks or Convolutional Neural Network (CNN) designs, which are common in contemporary machine learning research.  The absence of assessment on standardized benchmarks like as MNIST constrains the comparability of these findings to other research.  Subsequent research should investigate a broader range of problem types and network designs, together with benchmark datasets, to enhance the evaluation of the suggested methods' generalizability.  The research also failed to investigate a diverse range of metaheuristic variants.  Extensions include PSO with constriction factors, island models in GAs, and adaptive parameter tuning were not analyzed.  Emphasizing the gradient term in PSO-SGD during the latter stages of training may enhance convergence.  These improvements may result in more efficient training and necessitate additional inquiry.  Furthermore, although hyperparameters were optimized for particular configurations through several trials, their applicability to alternative settings is still ambiguous.  Due to constrained computational resources, comprehensive hyperparameter adjustment across configurations proved impractical, while it would be beneficial in subsequent research.

\section{Conclusion and Future Works}
\label{Conclusion and Future Works}

This study examined MHO strategies as viable alternatives to SGD for training ANNs on diverse regression tasks.  Our research demonstrates that under specified circumstances—especially in shallower or thinner networks with restricted computing iterations—metaheuristic optimization algorithms like PSO and GA can function as competitive or even superior alternatives to SGD.  Our findings indicate that the BBH may be applicable within the framework of GAs, particularly when crossover is utilized efficiently.  Additionally, we developed an innovative hybrid PSO-SGD algorithm that improves global search through gradient-based local optimization.  This method exhibited enhanced performance compared to regular PSO in confined environments, highlighting the efficacy of hybrid approaches.  These discoveries present potential opportunities for further investigation of hybrid and adaptive MHOs, especially in areas where conventional gradient-based approaches are inadequate due to problem complexity or computational constraints.

\section*{Declarations}

\begin{itemize}
    \item \textit{\textbf{A. Funding:}} No funds, grants, or other support was received.
    \item \textit{\textbf{B. Conflict of Interest:}} The authors declare that they have no known competing financial interests or personal relationships that could have appeared to influence the work reported in this paper.
    \item \textit{\textbf{C. Data Availability:}} Data will be made available on reasonable request.
    \item \textit{\textbf{D. Code Availability:}} Code will be made available on reasonable request.
\end{itemize}

\bibliographystyle{ieeetran}
\bibliography{main}

\end{document}